\documentclass[11pt]{article}
\usepackage{acl-hlt2011}
\usepackage{setspace}
\usepackage{times}
\usepackage{latexsym}
\usepackage{amsmath}
\usepackage{multirow}
\usepackage{url}

\usepackage[small,compact]{titlesec}

\title{Summarizing Decisions in Spoken Meetings}

\author{Lu Wang \\
  Department of Computer Science \\
  Cornell University \\
  Ithaca, NY 14853 \\
  {\tt luwang@cs.cornell.edu} \\\And
  Claire Cardie \\
  Department of Computer Science \\
  Cornell University \\
  Ithaca, NY 14853 \\
  {\tt cardie@cs.cornell.edu} \\}

\begin{document}

\maketitle

\vspace*{-0.5in}
\begin{abstract}
This paper addresses the problem of summarizing decisions in spoken meetings: our goal is to produce a concise  {\it decision abstract} for each meeting decision. We explore and compare token-level and dialogue act-level automatic summarization methods using both unsupervised and supervised learning frameworks. 
%
In the supervised summarization setting, and given true clusterings of decision-related utterances, 
we find that token-level summaries that employ discourse context can approach an upper bound 
for decision abstracts derived directly from dialogue acts.
In the unsupervised summarization setting,we find that summaries based on unsupervised partitioning 
of decision-related utterances perform comparably to those based on partitions generated using supervised 
techniques (0.22 ROUGE-F1 using LDA-based
topic models vs.\ 0.23 using SVMs). 
\end{abstract}

\section{Introduction}

Meetings are a common way for people to share information and discuss problems. And an effective meeting always leads to concrete decisions.
As a result, it would be useful to develop automatic methods that summarize not the entire meeting dialogue, but just the important decisions made. 
In particular, decision summaries would allow participants to review decisions from previous meetings as they prepare for an upcoming meeting. 
For those who did not participate in the earlier meetings, decision summaries might provide one type of efficient overview of the meeting contents.
For managers, decision summaries could act as a concise record of the idea generation process.

While there has been some previous work in summarizing meetings and conversations, very little work has focused on decision summarization: ~\newcite{Fernandez} and~\newcite{Bui:2009:EDM:1708376.1708410} investigate the use of a semantic parser and machine learning methods for phrase- and token-level decision summarization. We believe our work is the first to explore and compare token-level and dialogue act-level approaches --- 
using both unsupervised and supervised learning methods --- for summarizing decisions in meetings. 

\begin{table}[tbhp]
    {\footnotesize
    \setlength{\baselineskip}{0pt}
    \begin{tabular}{|l|}
        \hline
        {\bf C: Just spinning and not scrolling , I would say . (1)}\\
  C: But if you've got a [disfmarker] if if you've got a flipped \\thing , effectively it's something that's curved on one side\\ 
     and flat on the other side , but you folded it in half . (2)\\
  D: the case would be rubber and the the buttons , (3)\\
  {\bf B: I think the spinning wheel is definitely very now . (1)}\\
  B: and then make the colour of the main remote [vocal-\\ sound] the colour like vegetable colours , do you know ? (4)\\
  B: I mean I suppose vegetable colours would be orange\\ and green and some reds and um maybe purple (4)\\
  {\bf A: but since {\bf LCDs} seems to be uh a definite yes , (1)}\\
  A: Flat on the top . (2)\\~\\
{\bf Decision Abstracts (Summary)}\\
{\sc Decision 1}: The remote will have an LCD and spinning\\
     wheel inside.\\
{\sc Decision 2}: The case will be flat on top and curved on\\ the bottom.\\
{\sc Decision 3}: The remote control and its buttons will be\\ made of rubber.\\
{\sc Decision 4}: The remote will resemble a vegetable and\\ be in bright vegetable colors.\\
        \hline
    \end{tabular}
    }
    \caption{A clip of a meeting from the AMI meeting corpus~\cite{Carletta05theami}. A, B, C and D refer to distinct speakers; 
             the numbers in parentheses indicate the associated meeting decision: {\sc decision 1}, {\sc 2}, {\sc 3} or {\sc 4}. Also shown is the gold-standard (manual) abstract (summary) for each decision.}
\end{table}

Consider the sample dialogue snippet in Table 1, which is part of the AMI meeting corpus \cite{Carletta05theami}. The Table lists only {\it decision-related dialogue acts (DRDAs)} --- utterances associated with at least one decision made in the meeting.\footnote{These are similar, but not completely equivalent, to the {\it decision dialogue acts (DDAs)} of~\newcite{Bui:2009:EDM:1708376.1708410},~\newcite{Fernandez},~\newcite{Frampton:2009:RDD:1699648.1699659}.
The latter refer to all DAs that appear in a decision discussion even if they do NOT support any particular decision.}
The DRDAs are ordered by time; intervening utterances are not shown.
DRDAs are important because they contain critical information for decision summary construction. 
%

Table 1 clearly shows some challenges for
decision summarization for spoken meetings beyond the disfluencies, high word error rates, absence of punctuation, interruptions and  hesitations due to speech. First, different decisions can be discussed more or less concurrently; as a result, {\it the  utterances associated with a single decision are not contiguous in the dialogue}. In Table 1, the dialogue acts (henceforth, DAs) concerning {\sc Decision 1}, for example, are interleaved with DAs for other decisions. 
%
%
Second, {\it some decision-related DAs
contribute more than others to the associated decision}.  
In composing the summary for {\sc Decision 1}, for example, we might safely ignore the first DA for {\sc Decision 1}.
Finally, more so than for standard text summarization, {\it purely extract-based summaries are not likely to be easily interpretable}: DRDAs often contain text that is irrelevant to the decision and many will only be understandable if analyzed 
in the context of the surrounding utterances. 

In this paper, we study methods for decision summarization for spoken meetings. We assume that all decision-related DAs have been
identified 
and aim to produce a summary for the meeting in the form of concise {\it decision abstracts} (see Table 1), one for each decision made. 
In response to the challenges described above, we propose a summarization framework that includes:
\begin{small}
\begin{description}
  \setlength{\itemsep}{0pt}
    \setlength{\parsep}{0pt}
    \setlength{\topsep}{0pt}
    \setlength{\partopsep}{0pt}
    \setlength{\labelsep}{0.5em}
   \item [Clustering of decision-related DAs.] Here we aim to partition the decision-related utterances (DRDAs)
according to the decisions each supports. This step is similar in spirit to many standard text summarization techniques \cite{Salton:1997:ATS:256268.256282} that begin by grouping sentences according to semantic similarity. 

    \item [Summarization at the DA-level.] We select just the important DRDAs 
in each cluster. Our goal is to eliminate redundant and less informative utterances.  The selected DRDAs are 
then concatenated to form the decision summary.
    
    \item [Optional token-level summarization of the 
selected ] {\bf DRDAs.} Methods are employed to capture concisely the gist of each decision, discarding any distracting text. 

    \item [Incorporation of the discourse context as needed.]  We hypothesize that this will produce more interpretable summaries.

\end{description}
\end{small}

More specifically, we compare both unsupervised (TFIDF \cite{Salton:1997:ATS:256268.256282} and LDA topic modeling \cite{Blei:2003:LDA:944919.944937}) and (pairwise) supervised clustering procedures (using SVMs and MaxEnt) for partitioning DRDAs according to the decision each supports. We also investigate unsupervised methods and supervised learning for decision summarization at both the DA and token level, with and without the incorporation of discourse context.  
During training, the supervised decision summarizers are told which DRDAs for each decision are the most informative for constructing the decision abstract. 

Our experiments employ the aforementioned AMI meeting corpus: we
compare our decision summaries to the manually generated decision abstracts for each meeting and evaluate performance using the ROUGE-1 \cite{Lin:2003:AES:1073445.1073465} text summarization evaluation metric.  

In the supervised summarization setting, our experiments demonstrate that with true clusterings of decision-related DAs, token-level summaries that employ limited discourse context can approach an upper bound for summaries extracted directly from DRDAs\footnote{The upper bound measures the vocabulary overlap of each gold-standard decision summary with the complete text of all of its associated DRDAs.}  --- 0.4387 ROUGE-F1 vs.\ 0.5333.
When using system-generated DRDA clusterings, the DA-level summaries always dominate token-level methods in terms of performance.

For the  unsupervised summarization setting, we investigate the use of both unsupervised and supervised methods for the initial DRDA clustering step.  We find that summaries based on unsupervised clusterings perform comparably to those generated using supervised techniques (0.2214 ROUGE-F1 using LDA-based topic models vs.\ 0.2349 using SVMs). As in the supervised summarization setting, we observe that including additional discourse context boosts performance only for token-level summaries.

\section{Related Work}

There exists much previous research on automatic text summarization using corpus-based, knowledge-based or statistical methods~\cite{Maybury:1999:AAT:554275,Marcu:2000:TPD:517637}. 
Dialogue summarization methods, however, generally try to account for the special characteristics of speech. Among early work in this subarea,~\newcite{Zechner:2002:ASO:638178.638181} investigates speech summarization based on maximal marginal relevance (MMR) and cross-speaker linking of information. Popular supervised methods for summarizing speech --- including maximum entropy, conditional random fields (CRFs), and support vector machines (SVMs) --- are investigated in~\newcite{Buist04automaticsummarization},~\newcite{Xie08evaluatingthe} and~\newcite{Galley:2006:SCR:1610075.1610126}. Techniques for determining semantic similarity are used for selecting relevant utterances in~\newcite{Gurevych:2004:SSA:1220355.1220465}.

Studies in~\newcite{DBLP:conf/interact/BanerjeeRR05} show that decisions are considered to be one of the most important outputs of meetings. And in recent years, there has been much research on detecting decision-related DAs. ~\newcite{Hsueh:2007:ADD:1787422.1787442}, for example, propose maximum entropy classification techniques to identify DRDAs in meetings; ~\newcite{Fernandez:2008:MDD:1622064.1622095} develop a model of decision-making dialogue structure and detect decision DAs based on it; and ~\newcite{Frampton:2009:RDD:1699648.1699659} implement a real-time decision detection system.


~\newcite{Fernandez} and~\newcite{Bui:2009:EDM:1708376.1708410}, however, might be the most relevant previous work to ours. 
The systems in both papers run an open-domain semantic parser on meeting transcriptions to produce multiple short fragments, and then employ machine learning methods to select the phrases or words that comprise the decision summary. Although their task is also decision summarization, their gold-standard summaries consist of manually annotated words from the meeting while we judge performance using manually constructed decision abstracts as the gold standard.  The latter are more readable, but often use a vocabulary different from that of the associated decision-related utterances in the meeting.



Our work differs from all of the above in that we (1) incorporate a clustering step to partition DRDAs according to the decision each supports; (2) generate decision summaries at both the DA- and token-level; and (3) investigate the role of  discourse context for decision summarization.

In the following sections, we investigate methods for clustering DRDAs (Section 3) and generating DA-level and token-level decision summaries (Section 4).  In each case, we evaluate the methods using the AMI meeting corpus.

\section{Clustering Decision-Related Dialogue Acts}
We design a preprocessing step that facilitates decision summarization by clustering all of the decision-related dialogue acts according to the decision(s) it supports. Because it is not clear how many decisions are made in a meeting, we use a hierarchical agglomerative clustering algorithm (rather than techniques that require {\it a priori} knowledge of the number of clusters) and choose the proper stopping conditions.
In particular, we employ {\it average-link} methods: at each iteration, we merge the two clusters with the maximum average pairwise similarity among their DRDAs. In the following subsections, we introduce unsupervised and supervised methods for measuring the pairwise DRDA similarity.

\subsection{DRDA Similarity: Unsupervised Methods}

We consider two unsupervised similarity measures --- one based on the TF-IDF score from the Information Retrieval research community, and a second based on Latent Dirichlet Allocation topic models.

\paragraph{TF-IDF similarity.}
TF-IDF similarity metrics have worked well as a measure of document similarity.  As a result, we employ it as one metric for measuring the similarity of two DRDAs. Suppose there are ${L}$ distinct word types in the corpus. We treat each decision-related dialgue act ${DA_{i}}$ as a document, and represent it as an ${L}$-dimensional feature vector ${\overrightarrow{FV_{i}}=(x_{i1}, x_{i2}, ..., x_{iL})}$, where ${x_{ik}}$ is word ${w_{k}}$'s ${tf \cdot idf}$ score for ${DA_{i}}$. Then the (average-link) similarity of cluster ${C_{m}}$ and cluster ${C_{n}}$, $Sim\_TFIDF(C_{m}, C_{n})$, is defined as :\\$$\frac{1}{\mid C_{m} \mid \cdot \mid C_{n}\mid} \sum_{\substack{DA_{i}\in C_{m}\\ DA_{j}\in C_{n}}} \frac{\overrightarrow{FV_{i}}\cdot \overrightarrow{FV_{j}}}{\parallel \overrightarrow{FV_{i}} \parallel \parallel \overrightarrow{FV_{j}} \parallel} $$

\paragraph{LDA topic models.}
In recent years, topic models have become a popular technique for discovering the latent structure of ``topics" or ``concepts" in a corpus. Here we use the Latent Dirichlet Allocation (LDA) topic models of~\newcite{Blei:2003:LDA:944919.944937} --- unsupervised probabilistic generative models that estimate the properties of multinomial observations. In our setting, LDA-based topic models provide a soft clustering of the DRDAs according to the topics they discuss.\footnote{We 
cannot easily associate each topic with a decision because the number of decisions is not known {\it a priori}.} To determine the similarity of two DRDAs, we effectively measure the similarity of their term-based topic distributions.

To train an LDA-based topic model for our task\footnote{Parameter estimation and inference done by GibbsLDA++.}, we treat each DRDA as an individual document. After training, each DRDA, ${DA_{i}}$, is assigned a topic distribution ${\overrightarrow{\theta_{i}}}$ according to the learned model. Thus, we can define the similarity of cluster ${C_{m}}$ and cluster ${C_{n}}$, $Sim\_LDA(C_{m}, C_{n})$, as :\\
$$\frac{1}{\mid C_{m} \mid \cdot \mid C_{n}\mid} \sum_{\substack{DA_{i}\in C_{m}\\ DA_{j}\in C_{n}}} {{\overrightarrow{\theta_{i}}}\cdot {\overrightarrow{\theta_{j}}}} $$
\begin{table}
    \hspace{0.2cm}
    {\footnotesize
    \setlength{\baselineskip}{0pt}
    \begin{tabular}{|l|}
        \hline
        {\bf Features}\\ \hline
        number of overlapping words\\
        proportion of the number of overlapping words to the le-\\ngth of shorter DA\\
        TF-IDF similarity\\
        whether the DAs are in an adjacency pair (see 4.3)\\
        time difference of pairwise DAs\\
        relative dialogue position of pairwise DAs\\
        whether the two DAs have the same DA type\\
        number of overlapping words in the contexts (see 4.2)\\ \hline
        
    \end{tabular}
    }
    \caption{Features for Pairwise Supervised Clustering}
\end{table}
\begin{spacing}{0.9}
\begin{table*}
    \hspace{0.2cm}
    {\footnotesize
    \setlength{\baselineskip}{0pt}
    \begin{tabular}{|c|c|c|c|c|c|c|c|}
        \hline
\textbf{ }&\multicolumn{3}{|c|}{\textbf{B-cubed}}& \multicolumn{3}{|c|}{\textbf{Pairwise}}& \multicolumn{1}{|c|}{\textbf{VOI}}\\
\cline{3-8}
\hline
 &PRECISION&RECALL&F1&PRECISION&RECALL&F1& \\
  {\bf Baselines}& & & &  & & & \\
AllInOneGroup&  0.2854&  1.0000&  0.4441&  0.1823&  1.0000&  0.3083&  2.2279\\
ChoiSegment  &0.4235&  0.9657&  0.5888&  0.2390&  0.8493&  0.3730&  1.8061\\
  {\bf Unsupervised Methods}& & & &  & & & \\
TFIDF &  0.6840 & 0.6686&  0.6762&  0.3281&  0.3004&  0.3137&  1.6604\\            
LDA topic models &  0.8265&  0.6432&  {\bf 0.7235}&  0.4588&  0.2980&  {\bf 0.3613}&  {\bf 1.4203}\\
  {\bf Pairwise Supervised Methods}& & & &  & & & \\
SVM &  0.7593&  0.7466&  {\bf 0.7529}&  0.5474&  0.4821&  0.5127&  {\bf 1.2239}\\
MaxEnt&  0.6999&  0.7948&  0.7443&  0.4858  &0.5704  &{\bf 0.5247}  &1.2726\\

\hline
    \end{tabular}
    }
    \caption{Results for Clustering Decision-Related DAs According to the Decision Each Supports}
\end{table*}
\end{spacing}

\subsection{DRDA Similarity: Supervised Techniques}
In addition to unsupervised methods for clustering DRDAs, we also explore an approach based on {\it Pairwise Supervised Learning}: we develop a classifier that determines whether or not a pair of DRDAs supports the same decision. So each training and test example is a feature vector that is a function of two DRDAs: for $DA_{i}$ and $DA_{j}$, the feature vector is ${\overrightarrow{FV_{ij}}=f(DA_{i}, DA_{j})=\{fv^{1}_{ij}, fv^{2}_{ij},...,fv^{k}_{ij}\}}$. Table 2 gives a full list of features that are used. Because the annotations for the time information and dialogue type of DAs are available from the corpus, we employ features including time difference of pairwise DAs, relative position\footnote{Here is the definition for the relative position of pairwise DAs. Suppose there are $N$ DAs in one meeting ordered by time, $DA_{i}$ is the $i$th DA and $DA_{j}$ is positioned at $j$. So the relative position of $DA_{i}$ and $DA_{j}$ is $\frac{|i-j|}{N}$.} and whether they have the same DA type.

We employ Support Vector Machines (SVMs) and Maximum Entropy (MaxEnt) as our learning methods, because SVMs are shown to be effective in text categorization~\cite{citeulike:3340317} and MaxEnt has been applied in many natural language processing tasks~\cite{Berger:1996:MEA:234285.234289}. Given an ${\overrightarrow{FV_{ij}}}$, for SVMs, we utilize the decision value of ${{\mathbf w}^{T} \cdot \overrightarrow{FV_{ij}}+{\mathbf b}}$ as the similarity, where ${\mathbf w}$ is the weight vector and ${\mathbf b}$ is the bias. For MaxEnt, we make use of the probability of $P(SameDecision\mid \overrightarrow{FV_{ij}})$ as the similarity value.


\subsection{Experiments}

\paragraph{Corpus.}
We use the AMI meeting Corpus~\cite{Carletta05theami}, a freely available corpus of multi-party meetings that contains a wide range of annotations. 
The 129 scenario-driven meetings involve four participants playing different roles on a design team. A short (usually one-sentence) abstract is included that describes each decision, action, or problem discussed in the meeting; and each DA is linked to the abstracts it supports.
We use the manually constructed decision abstracts as gold-standard summaries and assume that all decision-related DAs have been identified (but not linked to the decision(s) it supports).

\paragraph{Baselines.}
Two clustering baselines are utilized for comparison. One baseline places all decision-related DAs for the meeting into a single partition ({\sc AllInOneGroup}). The second uses the text segmentation software of ~\newcite{Choi:2000:ADI:974305.974309} to partition the decision-related DAs (ordered according to time) into several topic-based groups ({\sc ChoiSegment}).

\paragraph{Experimental Setup and Evaluation.}
Results for pairwise supervised clustering were obtained using 3-fold cross-validation.
In the current work, stopping conditions for hierarchical agglomerative clustering are selected 
manually:
For the TF-IDF and topic model approaches, we stop when the similarity measure reaches 0.035 and 0.015, respectively;
For the SVM and MaxEnt versions, we use 0 and 0.45, respectively. We use the Mallet implementation for MaxEnt and the ${\textsc{SVM}^{light}}$ implementation of SVMs. 

Our evaluation metrics include ${b^{3}}$ (also called B-cubed)~\cite{citeulike:3363397}, which is a common measure employed in noun phrase coreference resolution research; 
a pairwise scorer that measures correctness for every pair of DRDAs; and a variation of information (VOI) scorer~\cite{Meila:2007:CCI:1232959.1233220}, which measures the difference between the distributions of the true clustering and system generated clustering. As space is limited, we refer the readers to the original papers for more details. For ${b^{3}}$ scorer and pairwise scorer, higher results represent better performance; for VOI, lower is better.\footnote{The MUC scorer is popular in coreference evaluation, but it is flawed in measuring the singleton clusters which is prevalent in the AMI corpus. So we do not use it in this work.}
\paragraph{Results.}
The results in Table 3 show first that all of the proposed clustering methods outperform the baselines. Among the unsupervised methods, the LDA topic modeling is preferred to TFIDF. For the supervised methods, SVMs
and MaxEnt produce comparable results.

\section{Decision Summarization}

In this section, we turn to decision summarization --- extracting a short description of each decision based on the decision-related DAs in each cluster. We investigate options for constructing an extract-based summary that consists of a single DRDA and an abstract-based summary comprised of keywords that describe the decision. 
For both types of summary, we employ standard techniques from text summarization, but also explore the use of dialogue-specific features and the use of discourse context.

\subsection{DA-Level Summarization Based on Unsupervised Methods}
We make use of two unsupervised methods to summarize the DRDAs in each ``decision cluster". The first method simply returns the longest DRDA in the cluster as the summary ({\sc Longest DA}). 
The second approach returns the decision cluster prototype, i.e., the DRDA with the largest TF-IDF similarity with the cluster centroid ({\sc Prototype DA}).
Although important decision-related information may be spread over multiple DRDAs, both unsupervised methods allow us to determine summary quality when summaries are restricted to a single utterance.

\subsection{DA-Level and Token-Level Summarization Using Supervised Learning}
Because the AMI corpus contains a decision abstract for each decision made in the meeting, we can
use this supervisory information to train classifiers that can identify informative DRDAs (for DA-level summaries)
or informative tokens (for token-level summaries).  
\begin{spacing}{0.9}
\begin{table}[!t]
    {\footnotesize
    \setlength{\baselineskip}{0pt}
    \begin{tabular}{|l|}
        \hline
        {\bf Lexical Features}\\ \hline
        unigram/bigram\\
    length of the DA\\
    contain digits?\\
    has overlapping words with next DA?\\
    next DA is a positive feedback?\\
    \hline
    
    {\bf Structural Features}\\ \hline
    relative position in the meeting?(beginning, ending, or else)\\
    in an AP?\\
    if in an AP, AP type\\
    if in an AP, the other part is decision-related?\\
    if in an AP, is the source part or target part?\\
    if in an AP and is source part, target is positive feedback?\\
    if in an AP and is target part, source is a question?\\
    \hline
    
    {\bf Discourse Features}\\ \hline
    relative position to ``WRAP UP" or ``RECAP"\\
    \hline
    {\bf Other Features}\\ \hline
    DA type\\
    speaker role\\
    topic\\
        \hline
    \end{tabular}
    }
    \caption{Features Used in DA-Level Summarization}
\end{table}
\end{spacing}

\begin{spacing}{0.9}
\begin{table}[!t]
    \hspace{0.6cm}
    {\footnotesize
    \setlength{\baselineskip}{0pt}
    \begin{tabular}{|l|}
        \hline
        {\bf Lexical Features}\\ \hline
        current token/current token and next token\\
    length of the DA\\
    is digit?\\
    appearing in next DA?\\
    next DA is a positive feedback?\\
    \hline
    
    {\bf Structural Features}\\ \hline
    see Table 3\\
    \hline
    
    {\bf Grammatical Features}\\ \hline
    part-of-speech\\
    phrase type (VP/NP/PP)\\
    dependency relations\\
    
    \hline
    {\bf Other Features}\\ \hline
    speaker role\\
    topic\\
    
        \hline
    \end{tabular}
    }
    \caption{Features Used in Token-Level Summarization}
\end{table}
\end{spacing}
\begin{spacing}{0.9}
\begin{table}
    \hspace{0.2cm}
    {\footnotesize
    \setlength{\baselineskip}{0pt}
    \begin{tabular}{|c|c|c|c|}
        \hline
  &{\bf PREC}&{\bf REC}&{\bf F1}\\

\hline

    {\bf True Clusterings}& & & \\
  Longest DA  &0.3655&  0.4077&  0.3545\\
  Prototype DA&0.3626&0.4140&    0.3539\\

  {\bf System Clusterings}& & & \\
  {\bf using LDA}& & & \\
  Longest DA  &0.3623&0.1892&    0.2214\\
  Prototype DA  &0.3669&  0.1887&  0.2212\\
  {\bf using SVMs}& & & \\
  Longest DA  &0.3719&  0.1261&  0.1682\\
  Prototype DA  &0.3816&  0.1264&  0.1700\\
    
  {\bf No Clustering}& & & \\
  Longest DA  &0.1039&  0.1382&  0.1080\\
  Prototype DA  &0.1350&0.1209  &  0.1138\\

\hline  
  
  {\bf Upper Bound} &0.8970  &0.4089  &{\bf 0.5333}\\
\hline
    \end{tabular}
    }
    \caption{Results for ROUGE-1: Decision Summary Generation Using Unsupervised Methods}
\end{table}
\end{spacing}

\vspace*{-0.1in}
\paragraph{Dialogue Act-based Summarization.}
Previous research (e.g., ~\newcite{Murray05extractivesummarization}, ~\newcite{Galley:2006:SCR:1610075.1610126}, ~\newcite{Gurevych:2004:SSA:1220355.1220465}) has shown that DRDA-level extractive summarization can be effective 
when viewed as a binary classification task. 
To implement this approach, we assume that the DRDA to be extracted for the summary is the one with the largest
vocabulary overlap with the cluster's gold-standard decision abstract.
This DA-level summarization method has an advantage that the summary maintains good readability without a natural language generation component.

\vspace*{-0.1in}
\paragraph{Token-based Summarization.}
As shown in Table 1, some decision-related DAs contain many useless words when compared with the gold-standard abstracts. 
As a result, we propose a method for token-level decision summarization that focuses on identifying critical keywords from the 
cluster's DRDAs.  We follow the method of~\newcite{Fernandez}, but use a larger set of features and different learning methods.

\vspace*{-0.1in}
\paragraph{Adding Discourse Context.}
For each of the supervised DA- and token-based summarization methods, we also investigate the role of the discourse context.
Specifically, we augment the DRDA clusterings with additional (not decision-related) DAs from the meeting dialogue: for each decision partition, we include the DA with the highest TF-IDF similarity with the centroid of the partition. 
We will investigate the possible effects of this additional context on summary quality. 

In the next subsection, we describe the features used for supervised learning of DA- and token-based decision summaries.

\subsection{Dialogue Cues for Decision Summarization}
Different from text, dialogues have some notable features that we expect to be useful for finding informative, 
decision-related utterances. 
This section describes some of the dialogue-based features employed in our classifiers.
The full lists of features are shown in Table 4 and Table 5.
\begin{spacing}{0.9}
\begin{table*}
    \hspace{1.5cm}
    {\footnotesize
    \setlength{\baselineskip}{0pt}
    \begin{tabular}{|c|c|c|c|c|c|c|}
        \hline
        \textbf{ }&\multicolumn{3}{|c|}{\textbf{CRFs}}& \multicolumn{3}{|c|}{\textbf{SVMs}}\\
        \cline{3-7}
    \hline
   &PRECISION&RECALL&F1&PRECISION&RECALL&F1\\

  \hline
    {\bf True Clusterings}& & & & & & \\

  DA &  0.3922&  0.4449&  {\bf 0.3789} &   0.3661&  0.4695&  0.3727\\ 
  Token&  0.5055&  0.2453&  0.3033 &  0.4953&  0.3788&  {\bf 0.3963}\\ 
  DA+Context&0.3753&  0.4372&  {\bf 0.3678}  &0.3595&  0.4449&  0.3640\\  
  Token+Context&0.5682&  0.2825&  0.3454&0.6213&  0.3868&  {\bf 0.4387}\\ 
  
\hline

  {\bf System Clusterings}& & & & & & \\    
  {\bf using LDA}& & & & & & \\   
  DA&0.3087&  0.1663&  {\bf 0.1935}&  0.3391&  0.2097&  {\bf 0.2349}\\ 
  Token&0.3379&  0.0911&  0.1307&  0.3760&  0.1427&  0.1843\\ 
  DA+Context&   0.3305&  0.1748&  {\bf 0.2041}&  0.2903&  0.1869&  {\bf 0.2068}\\ 
  Token+Context&  0.4557&  0.1198&  0.1727&0.4882&  0.1486&  0.2056\\
  

  {\bf using SVMs}& & & & & & \\   
  DA&  0.3508&  0.1884&  {\bf 0.2197}&  0.3592&  0.2026&  {\bf 0.2348}\\ 
  Token&  0.2807&  0.04968&  0.0777&  0.3607&  0.0885&  0.1246\\ 
  DA+Context& 0.3583&  0.1891&  {\bf 0.2221}&  0.3418&  0.1892&  {\bf 0.2213}\\ 
  Token+Context&  0.4891&  0.0822&  0.1288&  0.4873&0.0914&  0.1393\\ 
\hline  
  
  {\bf No Clustering}& & & & & & \\
     DA&  0.08673&  0.1957&  0.0993&  0.0707&  0.1979&  0.0916\\ 
    Token&  0.1906&  0.0625&  0.0868&  0.1890&  0.3068&  0.2057 \\

\hline
    \end{tabular}
    }
    \caption{Results for ROUGE-1: Summary Generation Using Supervised Learning}
\end{table*}
\end{spacing}

\vspace*{-0.05in}
\paragraph{Structural Information: Adjacency Pairs.}
An {\it Adjacency Pair} (AP) is an important conversational analysis concept; APs are considered the fundamental unit of conversational organization~\cite{schegloff1973ouc}. 
In the AMI corpus, an AP pair consists of a source utterance and a target utterance, produced by different speakers. The source precedes the target but they are not necessarily adjacent.
We include features to indicate whether or not two DAs are APs indicating {\sc question+answer} or {\sc positive feedback}. For these features, we use the gold-standard AP annotations. 
We also include one feature that checks membership in a small set of words to decide whether a DA contains positive feedback (e.g., ``yeah", ``yes").  

\vspace*{-0.05in}
\paragraph{Discourse Information: Review and Closing Indicator.}
Another pragmatic cue for dialogue discussion is terms like ``wrap up" or ``recap", indicating that speakers will review the key meeting content.  We include the distance between these indicators and DAs as a feature.

\vspace*{-0.05in}
\paragraph{Grammatical Information: Dependency Relation Between Words.}
For token-level summarization, we make use of the grammatical relationships in the DAs. As in~\newcite{Bui:2009:EDM:1708376.1708410} and~\newcite{Fernandez}, we design features that encode (a) basic predicate-argument structures involving major phrase types (S, VP, NP, and PP)
and (b) additional typed dependencies from~\newcite{citeulike:2830913}. We use the Stanford Parser.

\section{Experiments}

Experiments based on supervised learning are performed using 3-fold cross-validation. We train two different types of classifiers for identifying informative DAs or tokens: Conditional Random Fields (CRFs) (via Mallet) and Support Vector Machines (SVMs) (via ${\textsc{SVM}^{light}}$).

We remove function words from DAs before using them as the input of our systems. The AMI decision abstracts are the gold-standard summaries.
We use the ROUGE~\cite{Lin:2003:AES:1073445.1073465} evaluation measure. ROUGE is a recall-based method that can identify systems producing succinct and descriptive summaries.\footnote{We use the stemming option of the ROUGE software at \url{http://berouge.com/}.}

\vspace*{-0.05in}
\paragraph{Results and Analysis.}

Results for the unsupervised and supervised summarization methods are shown in Tables 6 and 7, respectively. In the tables, {\sc True clusterings} means that we apply our methods on the gold-standard DRDA clusterings. {\sc System clusterings} use clusterings obtained from the methods introduced in Section 4; we show results only using the best unsupervised ({\sc using LDA}) and supervised ({\sc using SVMs}) DRDA clustering techniques.

Both Table 6 and 7 show 
that some attempt to cluster DRDAs improves the summarization results vs.\ {\sc No Clustering}. In Table 6, there is no significant difference between the results obtained from the {\sc Longest DA} and {\sc Prototype DA} for any experiment setting. This is because the longest DA is often selected as the prototype. An {\sc Upper Bound} result is listed for comparison: for each decision cluster, this system selects all words from the DRDAs that are part of the decision abstract (discarding duplicates). 

Table 7 presents the results for supervised summarization. Rows starting with {\sc DA} or {\sc Token} indicate results at the DA- or token-level.
The {\sc +Context} rows show results when discourse context is included.\footnote{In our experiments, we choose the top 20 relevant DAs as context.}
%
We see that:
(1) SVMs have a superior or comparable summarization performance vs. CRFs on every task.
(2) 
Token-level summaries perform better than DA-level summaries only using {\sc True Clusterings} and the SVM-based summarizer.
(3) Discourse context generally improves token-level summaries but not DA-level summaries.\footnote{We do not extract words from the discourse context
and experiments where we tried this were unsuccessful.} 
(4) DRDA clusterings produced by (unsupervised) LDA lead to summaries that are quite comparable in quality to those generated from DRDA clusterings produced by SVMs (supervised). From Table 6, we see that F1 is 0.2214 when choosing longest DAs from LDA-generated clusterings, which is comparable with the F1s of 0.1935 and 0.2349, attained when employing CRF and SVMs on the same clusterings.

The results in Table 7 are achieved by comparing abstracts having function words with system-generated summaries without function words. To reduce the vocabulary difference as much as possible, we also ran experiments that remove function words from the gold-standard abstracts, but no significant difference is observed.\footnote{Given abstracts without function words, and using the clusterings generated by LDA and employ CRF on DA- and token-level summarization, we get F1s of 0.1954 and 0.1329, which is marginally better than the corresponding 0.1935 and 0.1307 in Table 7. Similarly, if SVMs are employed in the same cases, we get F1s of 0.2367 and 0.1861 instead of 0.2349 and 0.1843. All of the other results obtain negligible minor increases in F1.}

Finally, we considered comparing our systems to the earlier similar work of~\cite{Fernandez} and~\cite{Bui:2009:EDM:1708376.1708410}, but found that it would be quite difficult because they employ a different notion from DRDAs which is Decision Dialogue Acts(DDAs).
In addition, they manually annotate words from their DDAs as the gold-standard summary, guaranteeing that their decision summaries employ the same vocabulary as the DDAs.  We instead use the actual decision abstracts from the AMI corpus.

%
%
%
%
%
%
%
%
%
%
%

\begin{table}[tb]
    \hspace{0.1cm}
    {\footnotesize
    \setlength{\baselineskip}{0pt}
    \begin{tabular}{|l|}
        \hline

  {\bf DA (1)}: um of course , as [disfmarker] we , we've already \\talked about the personal face plates in this meeting , {\bf (a)}\\
  {\bf DA (2)}: and I'd like to stick to that . {\bf (a)}\\
  {\bf DA (3)}: Well , I guess plastic and coated in rubber . {\bf (b)}\\
  {\bf DA (4)}: So the actual remote would be hard plastic and \\the casings rubber . {\bf (b)}\\
  \hline

    {\bf Decision (a)}: Will use personal face plates.\\
  {\bf Decision (b)}: Case will be plastic and coated in rubber.\\    
  \hline
  {\bf Longest DA:}\\
  talked about personal face plates in meeting \\
  {\bf Prototype DA:}\\
  actual remote hard plastic casings rubber\\
  {\bf DA-level:}\\
  talked about personal face plates in meeting, like to \\stick to, guess plastic and coated in rubber, \\actual remote hard plastic casings rubber\\
  {\bf Token-level:}\\
  actual remote plastic casings rubber \\
  {\bf DA-level and Discourse Context:}\\
  talked about personal face plates in meeting, guess plastic\\ and coated in rubber, actual remote hard plastic casings\\ rubber\\
  {\bf Token-level and Discourse Context:}\\
  remote plastic rubber\\
  \hline
    \end{tabular}
    }
    \caption{Sample system outputs by different methods are in the third cell (methods' names are in bold). First cell contains four DAs. (a) or (b) refers to the decision that DA supports, which is listed in the second cell. }
\end{table}
\subsection{Sample Decision Summaries}

Here we show sample summaries produced using our methods (Table 8). We pick one of the clusterings generated by LDA consisting of four DAs which support two decisions and take SVMs as the supervised summarization method. We remove function words and special markers like ``[disfmarker]" from the DAs.


The outputs indicate that either the longest DA or prototype DA contains part of the decisions in this ``mixed" cluster. 
Adding discourse context refines the summaries at both the DA- and token-levels.

\section{Conclusion}
In this work, we explore methods for producing decision summaries from spoken meetings at both the DA-level and the token-level. We show that  clustering DRDAs before identifying informative content to extract can improve summarization quality. We also find that unsupervised clustering of DRDAs (using LDA-based topic models) can produce summaries of comparable quality to those generated from supervised DRDA clustering. Token-level summarization methods can be boosted by adding discourse context and outperform DA-level summarization when true DRDA clusterings are available; otherwise, DA-level summarization methods offer better performance. 


\begin{small}
\paragraph{Acknowledgments.} This work was supported in part by National Science Foundation Grants
IIS-0535099 and IIS-0968450, and by a gift from Google.
\end{small}
{\scriptsize
\bibliographystyle{acl}

\begin{thebibliography}{}

\bibitem[\protect\citename{Bagga and Baldwin}1998]{citeulike:3363397}
Amit Bagga and Breck Baldwin.
\newblock 1998.
\newblock {Algorithms for scoring coreference chains}.
\newblock In {\em In The First International Conference on Language Resources
  and Evaluation Workshop on Linguistics Coreference}, pages 563--566.

\bibitem[\protect\citename{Banerjee \bgroup et al.\egroup
  }2005]{DBLP:conf/interact/BanerjeeRR05}
Satanjeev Banerjee, Carolyn~Penstein Ros{\'e}, and Alexander~I. Rudnicky.
\newblock 2005.
\newblock The necessity of a meeting recording and playback system, and the
  benefit of topic-level annotations to meeting browsing.
\newblock In {\em INTERACT}, pages 643--656.

\bibitem[\protect\citename{Berger \bgroup et al.\egroup
  }1996]{Berger:1996:MEA:234285.234289}
Adam~L. Berger, Vincent J.~Della Pietra, and Stephen A.~Della Pietra.
\newblock 1996.
\newblock A maximum entropy approach to natural language processing.
\newblock {\em Comput. Linguist.}, 22:39--71, March.

\bibitem[\protect\citename{Blei \bgroup et al.\egroup
  }2003]{Blei:2003:LDA:944919.944937}
David~M. Blei, Andrew~Y. Ng, and Michael~I. Jordan.
\newblock 2003.
\newblock Latent dirichlet allocation.
\newblock {\em J. Mach. Learn. Res.}, 3:993--1022, March.

\bibitem[\protect\citename{Bui \bgroup et al.\egroup
  }2009]{Bui:2009:EDM:1708376.1708410}
Trung~H. Bui, Matthew Frampton, John Dowding, and Stanley Peters.
\newblock 2009.
\newblock Extracting decisions from multi-party dialogue using directed
  graphical models and semantic similarity.
\newblock In {\em Proceedings of the SIGDIAL 2009 Conference}, pages 235--243.

\bibitem[\protect\citename{Buist \bgroup et al.\egroup
  }2004]{Buist04automaticsummarization}
Anne~Hendrik Buist, Wessel Kraaij, and Stephan Raaijmakers.
\newblock 2004.
\newblock Automatic summarization of meeting data: A feasibility study.
\newblock In {\em in Proc. Meeting of Computational Linguistics in the
  Netherlands (CLIN}.

\bibitem[\protect\citename{Carletta \bgroup et al.\egroup
  }2005]{Carletta05theami}
Jean Carletta, Simone Ashby, Sebastien Bourban, Mike Flynn, Thomas Hain,
  Jaroslav Kadlec, Vasilis Karaiskos, Wessel Kraaij, Melissa Kronenthal,
  Guillaume Lathoud, Mike Lincoln, Agnes Lisowska, and Mccowan Wilfried
  Post~Dennis Reidsma.
\newblock 2005.
\newblock The ami meeting corpus: A pre-announcement.
\newblock In {\em In Proc. MLMI}, pages 28--39.

\bibitem[\protect\citename{Choi}2000]{Choi:2000:ADI:974305.974309}
Freddy Y.~Y. Choi.
\newblock 2000.
\newblock Advances in domain independent linear text segmentation.
\newblock In {\em Proceedings of the 1st North American chapter of the
  Association for Computational Linguistics conference}, pages 26--33.

\bibitem[\protect\citename{Fern\'{a}ndez \bgroup et al.\egroup
  }2008a]{Fernandez}
Raquel Fern\'{a}ndez, Matthew Frampton, John Dowding, Anish Adukuzhiyil,
  Patrick Ehlen, and Stanley Peters.
\newblock 2008a.
\newblock Identifying relevant phrases to summarize decisions in spoken
  meetings.
\newblock INTERSPEECH-2008, pages 78--81.

\bibitem[\protect\citename{Fern\'{a}ndez \bgroup et al.\egroup
  }2008b]{Fernandez:2008:MDD:1622064.1622095}
Raquel Fern\'{a}ndez, Matthew Frampton, Patrick Ehlen, Matthew Purver, and
  Stanley Peters.
\newblock 2008b.
\newblock Modelling and detecting decisions in multi-party dialogue.
\newblock In {\em Proceedings of the 9th SIGdial Workshop on Discourse and
  Dialogue}, pages 156--163.

\bibitem[\protect\citename{Frampton \bgroup et al.\egroup
  }2009]{Frampton:2009:RDD:1699648.1699659}
Matthew Frampton, Jia Huang, Trung~Huu Bui, and Stanley Peters.
\newblock 2009.
\newblock Real-time decision detection in multi-party dialogue.
\newblock In {\em Proceedings of the 2009 Conference on Empirical Methods in
  Natural Language Processing: Volume 3 - Volume 3}, pages 1133--1141.

\bibitem[\protect\citename{Galley}2006]{Galley:2006:SCR:1610075.1610126}
Michel Galley.
\newblock 2006.
\newblock A skip-chain conditional random field for ranking meeting utterances
  by importance.
\newblock In {\em Proceedings of the 2006 Conference on Empirical Methods in
  Natural Language Processing}, pages 364--372.

\bibitem[\protect\citename{Gurevych and
  Strube}2004]{Gurevych:2004:SSA:1220355.1220465}
Iryna Gurevych and Michael Strube.
\newblock 2004.
\newblock Semantic similarity applied to spoken dialogue summarization.
\newblock In {\em Proceedings of the 20th international conference on
  Computational Linguistics}.

\bibitem[\protect\citename{Hsueh and
  Moore}2008]{Hsueh:2007:ADD:1787422.1787442}
Pei-Yun Hsueh and Johanna~D. Moore.
\newblock 2008.
\newblock Automatic decision detection in meeting speech.
\newblock In {\em Proceedings of the 4th international conference on Machine
  learning for multimodal interaction}, pages 168--179.

\bibitem[\protect\citename{Joachims}1998]{citeulike:3340317}
Thorsten Joachims.
\newblock 1998.
\newblock {Text categorization with Support Vector Machines: Learning with many
  relevant features}.
\newblock In Claire N\'{e}dellec and C\'{e}line Rouveirol, editors, {\em
  Machine Learning: ECML-98}, volume 1398, chapter~19, pages 137--142.
  Berlin/Heidelberg.

\bibitem[\protect\citename{Lin and Hovy}2003]{Lin:2003:AES:1073445.1073465}
Chin-Yew Lin and Eduard Hovy.
\newblock 2003.
\newblock Automatic evaluation of summaries using n-gram co-occurrence
  statistics.
\newblock In {\em Proceedings of the 2003 Conference of the North American
  Chapter of the Association for Computational Linguistics on Human Language
  Technology - Volume 1}, pages 71--78.

\bibitem[\protect\citename{Mani}1999]{Maybury:1999:AAT:554275}
Inderjeet Mani.
\newblock 1999.
\newblock {\em Advances in Automatic Text Summarization}.
\newblock MIT Press, Cambridge, MA, USA.

\bibitem[\protect\citename{Marcu}2000]{Marcu:2000:TPD:517637}
Daniel Marcu.
\newblock 2000.
\newblock {\em The Theory and Practice of Discourse Parsing and Summarization}.
\newblock MIT Press, Cambridge, MA, USA.

\bibitem[\protect\citename{Marneffe \bgroup et al.\egroup
  }2006]{citeulike:2830913}
M.~Marneffe, B.~Maccartney, and C.~Manning.
\newblock 2006.
\newblock {Generating Typed Dependency Parses from Phrase Structure Parses}.
\newblock In {\em Proceedings of LREC-06}, pages 449--454.

\bibitem[\protect\citename{Meil\u{a}}2007]{Meila:2007:CCI:1232959.1233220}
Marina Meil\u{a}.
\newblock 2007.
\newblock Comparing clusterings---an information based distance.
\newblock {\em J. Multivar. Anal.}, 98:873--895, May.

\bibitem[\protect\citename{Murray \bgroup et al.\egroup
  }2005]{Murray05extractivesummarization}
Gabriel Murray, Steve Renals, and Jean Carletta.
\newblock 2005.
\newblock Extractive summarization of meeting recordings.
\newblock In {\em in Proceedings of the 9th European Conference on Speech
  Communication and Technology}, pages 593--596.

\bibitem[\protect\citename{Salton \bgroup et al.\egroup
  }1997]{Salton:1997:ATS:256268.256282}
Gerard Salton, Amit Singhal, Mandar Mitra, and Chris Buckley.
\newblock 1997.
\newblock Automatic text structuring and summarization.
\newblock {\em Inf. Process. Manage.}, 33:193--207, March.

\bibitem[\protect\citename{Schegloff and Sacks}1973]{schegloff1973ouc}
E.~A. Schegloff and H.~Sacks.
\newblock 1973.
\newblock {Opening up closings}.
\newblock {\em Semiotica}, 8(4):289--327.

\bibitem[\protect\citename{Xie \bgroup et al.\egroup }2008]{Xie08evaluatingthe}
Shasha Xie, Yang Liu, and Hui Lin.
\newblock 2008.
\newblock Evaluating the effectiveness of features and sampling in extractive
  meeting summarization.
\newblock In {\em in Proc. of IEEE Spoken Language Technology (SLT}.

\bibitem[\protect\citename{Zechner}2002]{Zechner:2002:ASO:638178.638181}
Klaus Zechner.
\newblock 2002.
\newblock Automatic summarization of open-domain multiparty dialogues in
  diverse genres.
\newblock {\em Comput. Linguist.}, 28:447--485, December.

\end{thebibliography}

}

\end{document}